# Evaluating Human Sensitivity to Gender Biased AI-Generated Content through Human Assessment of Google's Autocomplete Function

Tamara Serrano[1], Nikoleta Pantelidou[1], Raquel Montero[1], Natalia Moskvina[1], Paolo Morosi[1], Evelina Leivada[1,2]

[1.] Universitat Autònoma de Barcelona

[2.] Institució Catalana de Recerca i Estudis Avançats (ICREA)

Large Language Models (LLMs) are increasingly present, but their effects on human users remain underexplored. We tested whether AI-attributed information influences views regarding gender-related statements, a domain where biases are pervasive. Spanish-speaking participants recruited through Prolific evaluated biased statements derived from Google's autocomplete function and human-authored, non-biased gender-related facts. Participants were either informed that some of the material was AI-generated or received no information about its origin. AI attribution did not affect reaction times or agreement judgments. This result suggests that AI-generated content does not automatically raise credibility concerns and may be evaluated with less scrutiny compared to human-produced content.

## 1. Introduction

Artificial Intelligence (AI) is a transformative field that has revolutionized many scientific disciplines. As a simulation of human capacities, it enables machines to learn autonomously, making them increasingly convincing surrogates for human cognition. Large Language Models (LLMs), in particular, support tasks ranging from information retrieval to code-writing assistance, as well as original content generation and translation (Naveed et al., 2025). However, concerns have been voiced regarding their safety and credibility: these systems exhibit biases —systematic prejudices embedded in algorithms (Srinivasan et al., 2024)— that may produce skewed, inaccurate, or unfair outputs. Such biases largely originate in the training data, which, being predominantly human-generated, reflect societal biases and may reinforce harmful narratives and social inequalities (Abramski et al., 2025).

This study is situated within critical algorithm studies, a cross-disciplinary field spanning computer science, sociology, and communication, which examines bias in algorithmic systems (Díaz et al., 2018). Building on the "bias in, bias out" framework (Barocas & Selbst, 2016; Mayson, 2019), whereby outputs reflect biases in training data, we focus on Google's search engine, and specifically its autocomplete function. This feature relies on multiple implementation decisions, incorporating factors such as query frequency, geographical location, and user history (Miller & Record, 2017). Despite claims of neutrality and objectivity in search results (Purcell et al., 2012), prior work has found that AI-generated autocomplete forms have an influence on human cognition and behavior (Arnold et al., 2018; Jakesch et al., 2023), and that autocomplete suggestions may exert manipulative or harmful effects (Vicente & Matute, 2023; Sterling et al., 2026; Roy & Ayalon, 2020).

We test whether exposure to AI-attributed information affects human user responses. Specifically, we ask whether people respond differently when told that a given piece of information (biased or not) originates from AI. We focus on gender (i.e., statements about men and women); a domain that is particularly susceptible to biased representations (Lin et al., 2023). For example, when a search engine suggests "men need to have sexual relations" after receiving as input "men need" (retrieved autocomplete suggestion: "*El hombre necesita tener relaciones sexuales*"), does knowledge that such statements come from AI affect our degree of agreement or our processing speed, when we engage with this content? Our research questions are whether reaction times (RQ1) or offline judgments of agreement (RQ2) show a shift when participants are told that the content they engage with is potentially AI-generated. The experiment consisted of three types of questionnaires, administered online to 4 groups of participants across two rounds in a crossover design. Questionnaires A and B included 40 statements each: 20 gender-biased statements derived from Google's autocomplete function (e.g., "*El hombre reprime sus sentimientos*" ["Men suppress their feelings"]) and 20 human-authored statements describing well-established (i.e. neutral and non-biased) gender-related realities (e.g., "*Las mujeres suelen ocupar menos posiciones de poder*" ["Women tend to hold fewer positions of power"]). In questionnaire A, participants were not given any information about authorship, while in questionnaire B, authorship of some statements was attributed to AI. Under the AI attribution condition, participants did not know which statements were retrieved from Google; they were only told that some of the statements they will see were AI-generated. Questionnaire C, used as a control condition, comprised 40 human-authored statements. In this case, participants were told that these statements may contain outdated/old textbook beliefs about human traits (e.g., "*El frío es el principal causante del resfriado*" ["Cold weather is the main cause of having a cold"]). Two attention checks and two reliability items were included. We recruited 122 participants via Prolific, with a 5-day interval between the two rounds; each group was assigned a different experimental condition (see the repository for details).

## 2. Materials and Methods

All procedures adhered to the Declaration of Helsinki and were reviewed and approved by the Research Ethics Committee at the first author's host institute (ethics approval number CEEAH 7150). All participants provided informed consent before participating in the experiment.

The experiment consisted of administering three types of online questionnaires comprising 44 items (40 target statements, 2 attention-check items, and 2 reliability items) to four groups of participants (G1-G4) across two rounds (R1 and R2) separated by five days. Participants rated the perceived certainty of each statement on a Likert scale ranging from 1 (“Strongly disagree” [“*Totalmente en desacuerdo*”]) to 5 (“Strongly agree” [“*Totalmente de acuerdo*”]). Reaction times were recorded from the moment each statement appeared on screen until participants selected a response option.

Each experimental group initially included 30 participants (except G2, which had 32). Six participants did not complete the second round, resulting in a total of 122 participants in Round 1 and 116 in Round 2. Three different questionnaires were constructed. Two of them (A and B) included 20 statements extracted from Google’s autocomplete function and 20 human-authored statements reflecting non-biased gender-related realities. The third questionnaire (C) consisted of 40 statements reflecting outdated beliefs across different domains of human knowledge. These questionnaires were administered to the four groups as shown in Table 1.

Each questionnaire corresponded to one experimental condition (see Table 1): participants who received no information about statement origin, participants informed that some statements originated from AI, and participants informed about the origin of the false-belief statements. Participants were recruited via Prolific and received £2.25 for Round 1 and £2.75 for Round 2. We recruited native Spanish speakers regardless of geographical background. The final sample consisted of 89 men and 32 women aged between 19 and 65 years, plus one person who defined themselves as “Other”. For further details regarding the experimental design, data collection, and data cleaning procedures, see the repository at https://github.com/Tam1813/Tam1813-Evaluating-Human-Sensitivity-to-Gender-Biased-AI-Generated-Content-. The dataset, the code to run the analysis and the prompts are also available there.

## 3. Results

Data analysis was conducted in three stages. First, we considered the overall distribution of data across questionnaires, conditions, and rounds, without excluding neutral responses (i.e. 3 in the employed 1-5 Likert scale). Second, we performed a more fine-grained analysis focusing on the AI-generated and human-authored statement blocks, examining them according to condition and round, again without excluding neutral responses. Finally, the analyses were repeated after excluding neutral responses, yielding no substantial changes in the results (see the repository for a detailed account).

First, considering overall reaction time (RT) patterns, the distributions were positively skewed and exhibited high kurtosis, violating normality assumptions required for parametric tests. We therefore relied exclusively on nonparametric analyses. Effect sizes were quantified using $\eta^2$ for the Kruskal-Wallis test and $r = |Z| / \sqrt{N}$ for pairwise comparisons (Mann-Whitney and Wilcoxon), following conventional thresholds for small ($r \geq 0.10$), medium ($r \geq 0.30$), and large ($r \geq 0.50$) effects.

The Kruskal-Wallis test revealed a significant overall difference across the three questionnaires, $H(2) = 98.93$, $p < 0.001$, $\eta^2 = 0.012$, N = 7,857. However, the effect size was small: condition accounted for only 1.2% of total RT variance. Post hoc Mann-Whitney tests with Bonferroni correction ($\alpha_corr = 0.017$) showed that A and B differed significantly (U = 4,329,290, Z = -2.522, p = 0.012, p_Bonf = 0.035, r = 0.033), although the effect was trivial. Comparisons between A and C (U = 2,325,124, Z = -9.586, p_Bonf < 0.001, r = 0.138) and between B and C (U = 2,423,740, Z = -7.769, p_Bonf < 0.001, r = 0.111) were highly significant, albeit small in magnitude. Within-subject comparisons were conducted using Wilcoxon tests on median RTs per participant and questionnaire. In G(roup) 1, median RT increased from 2,816 ms to 3,089 ms (W = 143, Z = -1.366, p = 0.178, r = 0.258). In G2, median RT decreased from 3,053 ms to 2,909 ms (W = 224, Z = −0.175, p = 0.871, r = 0.032). Figure 1, Panel A shows these results.

Second, regarding offline judgments, questionnaires A and B (i.e., no-information vs. AI attribution conditions, respectively) showed nearly identical distributions: median = 4.0 and mean = 3.6 in both cases, with similar proportions of agreement (≥ 4), neutral (= 3), and disagreement (≤ 2) responses. A Kruskal-Wallis test detected significant global differences, largely driven by the divergence of questionnaire C (median = 2.0). The comparison between questionnaires A and B (Figure 2, Panel A) yielded identical medians and a null effect size, both between and within subjects (Wilcoxon in G1 and G2).

We examined the effects of gender and age on the dependent variables, both agreement judgments and RTs, using Mann-Whitney tests and Kruskal-Wallis tests followed by Dunn's post hoc tests with Bonferroni correction (4 age groups: 18-25, 26-35, 36-45, and 46+). RTs showed a modest increase with age. No consistent effects of gender or age were observed for agreement judgments.

Third, a 2×2 factorial analysis crossing actual statement source (items A1-A20 and B1-B20, AI-generated; items A21-A40 and B21-B40, human-authored) with experimental condition (no-information vs. AI-attributed information) revealed several effects. Within questionnaire A, where participants did not receive any information about source of authorship, RTs did not differ between blocks, either in between-subjects analyses (Mann-Whitney: Z = -0.174, p = 0.862,

r = 0.003) or within-subjects analyses (Wilcoxon: W = 1,936, Z = -0.449, p = 0.654, r = 0.047), with nearly identical medians (3,094 ms vs. 3,065 ms). In contrast, agreement judgments showed a large effect: AI-generated statements received neutral ratings (median = 3.0), whereas the remaining (human-authored statements) were associated with higher agreement (median = 4.0). This was the largest effect (Figures 1 and 2, Panel B), both between subjects (Mann-Whitney: Z = -24.483, p < 0.001, r = 0.448) and within subjects (Wilcoxon: W = 0, Z = -8.239, p < 0.001, r = 0.868).

Regarding condition effects, within-subject analyses in G1 and G2 (n = 58 pairs) showed no significant effect of AI attribution on the processing of AI-generated statements, either in RTs (W = 734, Z = -0.941, p = 0.347, r = 0.124; medians: 2,929 ms → 2,970 ms) or in agreement judgments (p = 0.192; medians: 3.0 → 3.0). Group-specific analyses confirmed this pattern (G1: p_RT = 0.902, r = 0.026; G2: p_RT = 0.198, r = 0.238). Judgments did not change, with the median remaining at 3.0. By contrast, the AI condition had a significant effect on human-generated items: participants required 309 ms longer to process these items when they were told that these items may have come from AI agents (median: 2,909 ms → 3,218 ms; Wilcoxon G1+G2: W = 528, Z = -2.536, p = 0.011, r = 0.333). The effect was significant only in G1 (W = 108, Z = -2.163, p = 0.031, r = 0.409), where the condition was presented in the second round, but not in G2, where it appeared in the first (p = 0.140, r = 0.272). Agreement for the human-authored items increased (median: 4.0 → 5.0), although this did not reach significance, despite a sizeable effect. For questionnaire C, items yielded significantly higher RTs than those in questionnaires A (Mann-Whitney: Z = 8.081, p < 0.001, r = 0.140; medians: 3,546 ms vs. 3,065 ms) and B (Z = 4.412, p < 0.001, r = 0.077; medians: 3,546 ms vs. 3,345 ms). The median for judgments of agreement was 2.0 in questionnaire C, compared to 4.0 for the human-authored items in questionnaire A and 5.0 in questionnaire B (Figure 2).

## 4. Discussion

Until recently, if one put in the search engine of Google the prompt “things that start with U” in Spanish, one of the items in the list was “*una mujer*” (“a woman”). This result is surprising because the retrieved suggestion woman is not a thing. A multimodal perspective confirms this pattern: googling images of occupations has been found to amplify gender biases in participants’ beliefs and hiring preferences (Guilbeault et al., 2023). More broadly, LLMs and other AI systems that rely on the same token-predicting technology have been linked to a tendency to amplify misinformation and project different types of biases (Kidd & Birhane, 2023).

When algorithmic predictions present gender, racial, and ethnic stereotypes that undermine or misrepresent a group, they enact a biased position which contributes to spreading misinformation about certain groups. How do humans engage with biased content, depending on its source, human- vs. AI-authored? Holding LLMs accountable to a different standard than humans raises concerns about the standards of testing the field wants to adopt on the road to ethical AI (Leivada et al., 2024). The present study taps into the accountability standards, positing two RQs: whether reaction times (RQ1) and offline judgments (RQ2) change based on whether participants are told that the statements they engage with may originate from AI.

Our results suggest that telling people that a statement comes from AI does not significantly affect either processing time or degree of (dis)agreement. Despite previous research showing the potential of AI agents to modulate human cognition and behavior (Arnold et al., 2018; Jakesch et al., 2023), our study suggests that humans are, in fact, only minimally influenced by explicit AI-authorship attribution. In this sense, a global comparison of the questionnaires reveals, first, that judgments remain stable across questionnaires A and B, with no significant shift depending on authorship attribution. This pattern persists even when neutral responses (Likert = 3) are excluded from the analysis. Second, concerning possible changes in processing time across questionnaires, we observed that the intraindividual increase in RTs was marginal and non-significant. Once again, this pattern remained stable when neutral, mid-range responses were excluded from the analysis. Overall, these findings suggest that participants were not significantly influenced when told that they were exposed to information potentially generated by AI.

However, when a more fine-grained comparison is conducted focusing on the blocks of statements included in questionnaires A and B (i.e., AI-generated, biased vs. human-authored statements describing well-established gender-related patterns), more specific patterns emerge. First, when participants are not provided with any information regarding the origin of the statements, they show higher agreement rates when engaging with the non-biased gender-related statements. In this regard, one could argue that either there exists some form of implicit detection of the linguistic style characteristic of AI that is a priori in place, leading to largely uniform performance across our two critical conditions, AI attribution or absence of it, or that participants perhaps question the biased statements themselves, which would explain their greater agreement with the non-biased gender-related items. These findings add a different angle to the literature that reveals a robust impact of AI on human behavior and decision-making, even in the long term (Vicente & Matute, 2023).

When focusing on the effect of condition within each group and round, a striking pattern can be observed. When participants start the first round of the experiment without receiving any information about the origin of the statements (no-information condition), both RTs and the degree of agreement with non-biased gender-related statements are higher than those observed for AI-generated ones. However, this effect is only significant in G1, and not in G2, the group that initially received information regarding the origin of some of the statements (i.e. the AI attribution condition). This finding suggests that the mental framework derived from knowledge about the source of information carries over into the second questionnaire, even when no information about authorship is subsequently provided. This finding could be interpreted as entailing a certain degree of distrust toward AI, a greater level of reflective processing, and, by contrast, a more positive evaluation of human-generated content. Considering previous research claiming that AI can distort beliefs and contribute to the emergence of biases, this finding is interesting, as it points to a certain degree of awareness (or even resistance) on the part of human participants.

Regarding the effects of demographic variables, it is noteworthy that we did not find evidence indicating that either gender or age influenced participants' responses to the statements, whether AI-generated or human-generated. Given that these statements contained strongly biased claims concerning men and women, one might have expected either slower processing or shifts in the degree of agreement associated with such statements, particularly with respect to the gender variable, in light of the continued salience of gender issues in contemporary society. This finding raises the question of whether we may, in fact, be less aware than we assume of the ways in which AI contributes to the perpetuation of gender-related biases (Lin et al., 2023).

As highlighted throughout this study, AI systems exhibit several biases that have increasingly become the focus of research in recent years (Kidd & Birhane, 2023). The impact of certain AI tools, particularly applications based on LLMs, has also attracted considerable attention as it affects human behavior and cognitive capacities, including attitudes, decision-making, and belief systems (Arnold et al., 2018, Jakesch et al., 2023, Sterling et al., 2026). Nevertheless, our findings revealed that neither RTs (RQ1) nor offline judgments of agreement (RQ2) are globally affected by explicit knowledge of exposure to AI-generated information. This leads us to consider the possible degree of tolerance toward the perpetuation of biases already present within society itself. In this regard, the fact that the AI attribution condition operated in the manner previously described — namely, producing a significant increase in RTs and a shift in judgments concerning well-established, non-biased gender-related statements only in G1 (no-information start)— may suggest the existence of a certain skepticism toward AI-generated content; one that does not disappear under the mere labeling of this condition. What does change, and only for human-generated statements, is that awareness of the informational context (in this case, knowing that some content originates from AI) leads to an even more positive evaluation of human-authored content, possibly by contrast or, perhaps more optimistically, because participants display some degree of reservation, however small, toward certain forms of biased information.

Regarding the limitations of the present study, in terms of drawing conclusions that may be generalized to other AI systems, one aspect that could be further examined in future work concerns the topic of assessment, amplifying the range of tested biases but also more tightly controlling the testing stimuli used. The differences between questionnaires A and B, particularly in the nature of the statements they contained, despite our attempt to maximize similarity between them (see the repository for further details), entail that part of the observed effect may stem from variability in the intrinsic linguistic characteristics of the statements themselves, such as exact length or lexical choices. Last, regarding demographic data, future studies could use samples with more balanced distributions of age and gender across all groups.

## Figures

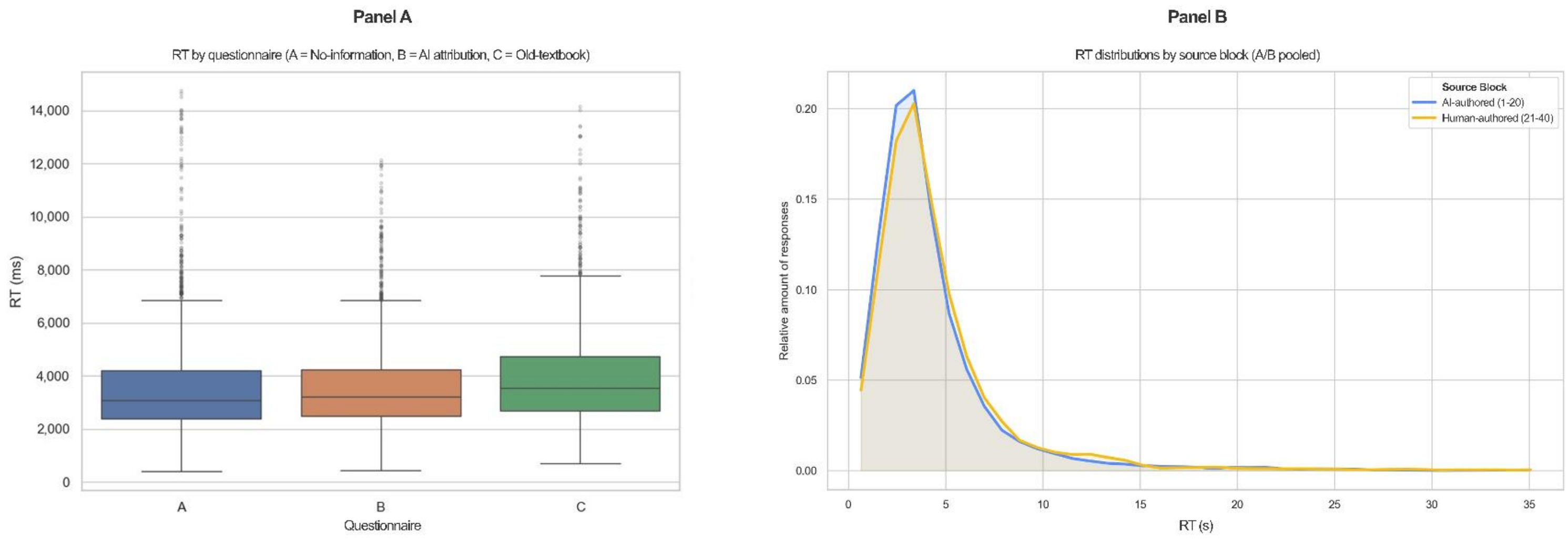

**Figure 1**. RT distribution: Full analysis (ms) and AI-generated vs. human-authored contrast (s)

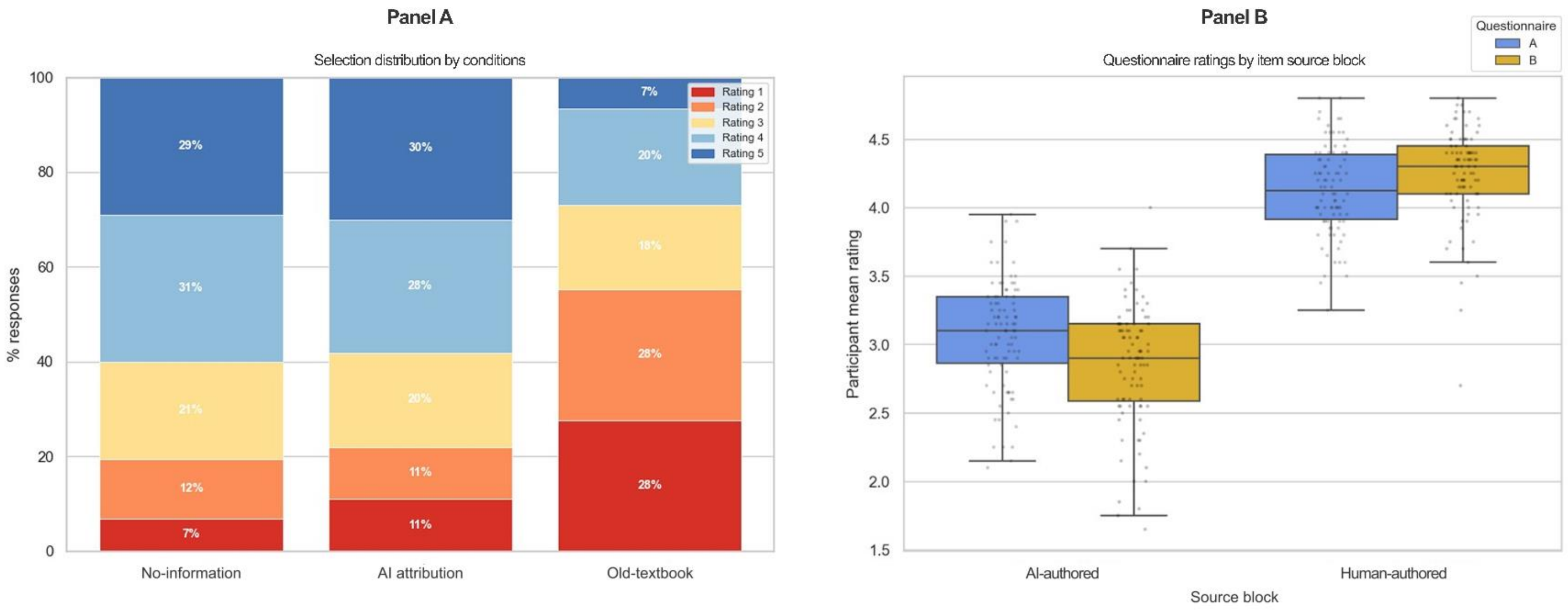

**Figure 2**. Judgment distributions on a 5-point Likert scale: Full analysis and AI-generated vs. human-authored contrast

**Tables**

**Table 1**. Questionnaire and Condition Distribution by Groups and Rounds

| | Round 1 | Round 2 |
|---|---|---|
| **G1** | A → No-information condition | B → AI attribution condition |
| **G2** | B → AI attribution condition | A → No-information condition |
| **G3** | A → No-information condition | C → Old-textbook attribution condition |
| **G4** | C → Old-textbook attribution condition | B → AI attribution condition |